\documentclass[sn-basic]{sn-jnl}

\usepackage[T1]{fontenc}
\usepackage[retain-explicit-plus]{siunitx}
\usepackage{xcolor}
\usepackage{booktabs}
\usepackage{tabu}
\usepackage{tabularx}
\usepackage{tabulary}
\usepackage{multirow,bigdelim}
\usepackage{rotating}
\usepackage{makecell}
\usepackage{tablefootnote}
\usepackage{threeparttable}
\usepackage{listings}
\usepackage{enumitem}
\usepackage{comment}
\usepackage{tablefootnote}
\usepackage{subcaption}
\usepackage{enumitem,amssymb}
\usepackage{graphicx}
\usepackage{makecell}
\usepackage{lineno}
\usepackage{multirow}

\definecolor{yellow}{rgb}{0.99, 0.76, 0.0}

\usepackage{blindtext}
\usepackage{hyperref}
\usepackage{nameref}

\newcounter{mylabelcounter}

\makeatletter
\newcommand{\labelText}[2]{%
\refstepcounter{mylabelcounter}%
\immediate\write\@auxout{%
 \string\newlabel{#2}{{\unexpanded{#1}}{\thepage}{{\unexpanded{#1}}}{mylabelcounter.\number\value{mylabelcounter}}{}}%
}%
}
\makeatother
\definecolor{green}{rgb}{0.01, 0.75, 0.24}

\newlist{todolist}{itemize}{2}
\setlist[todolist]{label=$\square$}

\jyear{2022}

\makeatletter
\newcommand\footnoteref[1]{\protected@xdef\@thefnmark{\ref{#1}}\@footnotemark}
\makeatother

\begin{document}

This version of the article has been accepted for publication, after peer review but is not the Version of Record and does not reflect post-acceptance improvements, or any corrections. The Version of Record is available online at: https://doi.org/10.1007/s11191-024-00530-2.

\title[A review on the use of LLMs as virtual tutors]{A review on the use of large language models as virtual tutors}

\author*[1]{\fnm{Silvia} \sur{García-Méndez}}\email{sgarcia@gti.uvigo.es}

\author[1]{\fnm{Francisco} \sur{de Arriba-Pérez}}\email{farriba@gti.uvigo.es}

\author[2]{\fnm{María del Carmen} \sur{Somoza-López}}\email{carmensomoza@uvigo.gal}

\affil*[1]{Information Technologies Group, atlanTTic, University of Vigo, Vigo, Spain}
\affil*[2]{Applied Mathematics I Department, University of Vigo, Spain}

\abstract{Transformer architectures contribute to managing long-term dependencies for Natural Language Processing, representing one of the most recent changes in the field. These architectures are the basis of the innovative, cutting-edge Large Language Models (\textsc{llm}s) that have produced a huge buzz in several fields and industrial sectors, among the ones education stands out. Accordingly, these generative Artificial Intelligence-based solutions have directed the change in techniques and the evolution in educational methods and contents, along with network infrastructure, towards high-quality learning. Given the popularity of \textsc{llm}s, this review seeks to provide a comprehensive overview of those solutions designed specifically to generate and evaluate educational materials and which involve students and teachers in their design or experimental plan. To the best of our knowledge, this is the first review of educational applications (\textit{e.g.}, student assessment) of \textsc{llm}s. As expected, the most common role of these systems is as virtual tutors for automatic question generation. Moreover, the most popular models are \textsc{gpt-3} and \textsc{bert}. However, due to the continuous launch of new generative models, new works are expected to be published shortly.}

\keywords{Artificial Intelligence, assessment, Large Language Models, review, virtual tutor}

\maketitle

\section{Introduction}

Artificial Intelligence (\textsc{ai}) refers to the synthetic capabilities of computer science applications to perform tasks that usually require human intelligence (\textit{e.g.}, adaptation, learning, reasoning, etc.) \citep{Sarker2022,Cooper2023}. The recent technological advancements within the \textsc{ai} field have led to relevant changes in business and research, the economy, and society (\textit{i.e.}, mega-trends) that are predicted to continue \citep{Estigarribia2022,Haluza2023,Rasa2023}. 

The most relevant change that perfectly exemplifies the impact of the mega-trends above is the transformer architectures that contribute to managing long-term dependencies for Natural Language Processing (\textsc{nlp}) \citep{Tay2023}. They are the basis of the innovative, cutting-edge Large Language Models (\textsc{llm}s) that have produced a huge buzz in several fields and industrial sectors \citep{MacNeil2022}. In this line, Chat\textsc{gpt} achieved more than 1 million users within the first five days of its release\footnote{Available at \url{https://twitter.com/gdb/status/1599683104142430208}, April 2024.}. Accordingly, \textsc{llm}s have been used in economy and finance \citep{Alshater2022}, journalism \citep{Pavlik2023}, medicine \citep{Siobhan2023}, and education \citep{Sallam2023}, among others. However, as other technological advancements, \textsc{llm}s have experienced the community's resistance, a common evolutionary and social psychology phenomenon \citep{Tobore2019}.

Regarding the learning field, during the 21st century, education has experienced a profound change in methods and content. Specifically, a flexible and multidisciplinary environment is sought, where the student can actively participate in their learning process, promoting more autonomous and ubiquitous studying thanks to \textsc{ai} advancements \citep{Baidoo2023,Li2023}. The scientific community has researched the use of \textsc{ai} techniques like Machine Learning (\textsc{ml}) models for training purposes ever since their inception \citep{Hochberg2018,talan2021artificial,Huang2022}, causing progressive advances towards autonomous high-quality learning \citep{Han2018,Demircioglu2022}. Mainly, \textsc{ai} has driven the technological transition in this field regarding the instructional applications, contents, platforms, resources, techniques, tools, and network infrastructure \citep{Roll2016}. This transition also involves changes in the leading roles of the education systems, teachers, and students since this new digital education environment requires new digital competencies and reasoning patterns \citep{Jensen2018,Zhou2023}. However, although promising, the advances offered by \textsc{ai} are still far from becoming a standardized tool in the educational field due to its early state and the need for training in using these solutions to take the most advantage of them.

Of particular interest is the impact of those applications that leverage \textsc{llm}s, framed within the generative \textsc{ai} field and based on \textsc{ml} techniques. They enable hands-on learning and are common practice in the classroom nowadays. Compared to previous \textsc{ai} solutions and traditional methodologies, which focused primarily on modifying the textual input using correction, paraphrasing, and sentence completion techniques, \textsc{llm}s generate on-the-fly human-like utterances, hence its popularity, especially among students and teachers \citep{Rudolph2023}. Current advanced \textsc{llm}s can enhance pedagogical practice and provide personalized assessment and tutoring \citep{Sok2023}. Consideration should be given to the cooperation between \textsc{llm}s-based systems and humans, provided the experience and scientific knowledge along with the capabilities of the human-agents for creativity and emotional intelligence \citep{Zhang2020,Korteling2021}. Note that these \textsc{ai}-based systems present advantages in specific educational tasks as self-learning tools and virtual tutors. Specifically, they enable automatic answer grading \citep{ahmed2022application}, explanation generation \citep{humphry2023potential}, question generation \citep{bhat2022towards}, and problem resolution \citep{zong2022solving}. Furthermore, when used for text summarization \citep{phillips2022exploring}, they help synthesize content and improve the student's abstraction capabilities. Their use as learning software in virtual assistants is highly relevant to flexible learning \citep{designing_wang2022,yamaoka2022experience}. Furthermore, their language intelligence capabilities make them an appropriate tool for code correction \citep{macneil2022generating}.

More in detail, \textsc{llm}s are trained with massive textual data sets to create human-like utterances. They perform a wide variety of \textsc{nlp} taking advantage of fine-tuning and pre-training pipelines \citep{Kasneci2023}. Note the relevance of both the pre-training and prompt engineering development. The first concept refers to training \textsc{llm}s with miscellaneous large data sets, while the second refers to specific fine-tuning on a particular task \citep{Kasneci2023}. Consequently, the quality of the \textsc{llm}s output highly depends on the input data and prompt designed, aka prompt engineering \citep{Cooper2023}. The latter technique ranges from zero-shot learning, widely popular when applied to \textsc{llm}s \citep{Russe2024}, to few-shot learning. Note that the model follows task instructions in zero-shot learning since the end user provides no examples. In contrast, in few-shot learning, the model learns from the demonstrations available (\textit{i.e.}, few-shot text prompts).

Among the most popular \textsc{llm}s, \textsc{bert} \citep{Devlin2019}, \textsc{gpt-3} \citep{gpt3-2020}, \textsc{gpt-3.5}\footnote{Available at \url{https://platform.openai.com/docs/models/gpt-3-5}, as referred to the \texttt{text-davinci-003} and \textsc{gpt-3.5}-turbo models, April 2024.}, \textsc{gpt-4}\footnote{Available at \url{https://platform.openai.com/docs/models/gpt-4}, April 2024.} and \textsc{t5} \citep{Raffel2020} deserve attention. \textsc{bert} (Bidirectional Encoder Representations from Transformers) was released by Google in October 2018 (slightly after \textsc{gpt-1} dated June 2018). It is a pre-trained transformer-based encoder model that can be fine-tuned on specific \textsc{nlp} tasks such as Named Entity Recognition (\textsc{ner}), question answering, and sentence classification. Moreover, \textsc{gpt-3}, \textsc{gpt-3.5}, and \textsc{gpt-4} (Generative Pre-trained Transformer) models were released by OpenAI in 2020, 2022 and 2023, respectively. More in detail, \textsc{gpt-4} is already deployed in Chat\textsc{gpt} application, which compared to other \textsc{llm}s can generate context-aligned responses and interact naturally with the end users as a peer. This model goes beyond producing reports and translating assessments by creating source code \citep{Haleem2022} and responding to complex questions posed by the students in real-time \citep{Shaji2023}. It can also show creativity to some extent in writing \citep{Baidoo2023}. \textsc{t5} (Text-to-text Transfer Transformer) model was released by Google following the encoder-decoder transformer architecture in 2020. Even though its configuration is similar to \textsc{bert}, it differs in some steps of the pipeline, like pre-normalization \citep{Pipalia2020}.

Given the widespread of \textsc{ai}-based solutions in our everyday lives and particularly the popularity of advanced \textsc{nlp}-based chatbots for learning purposes to generate and evaluate educational materials, this review seeks to provide a comprehensive overview of the systems that exploit \textsc{llm}s and were explicitly designed for educational purposes (\textit{i.e.}, virtual tutors for question generation and assessment). Thus, involving students or teachers at the design or evaluation levels, excluding those works in which the application of the solution for educational use cases was feasible but not initially designed for that purpose. The ultimate objective is to promote the advancement of these existing solutions in a collaborative environment between academia (\textit{i.e.}, researchers and developers) and end-users (\textit{i.e.}, students and teachers). To the best of our knowledge, this is the first review in this regard. Note that there exist few review works that focused on specific related fields such as health care education \citep{Sallam2023,Sallam2023} or specific features like the responsible and ethical use of \textsc{llm}s \citep{Mhlanga2023} and their impact on academic integrity \citep{Perkins2023}.

The rest of this paper is organized as follows. Section \ref{sec:methodology} describes the methods and materials used in the review. Section \ref{sec:discussion} presents the discussion on the selected relevant works. Finally, Section \ref{sec:conclusion} concludes the article and details future research.

\section{Methodology}
\label{sec:methodology}

\begin{figure}[!htbp]
\centering
\includegraphics[width=1\textwidth]{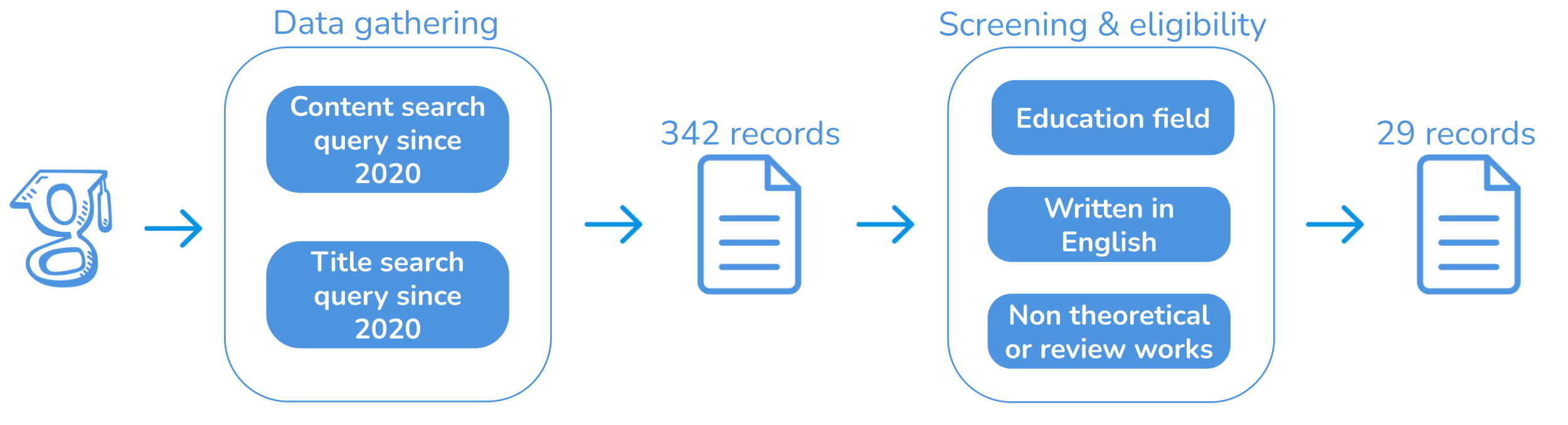}
\caption{\label{fig:scheme}Review pipeline.}
\end{figure}

The review methodology followed is composed of two steps: (\textit{i}) data gathering (Section \ref{sec:data_gathering}), and (\textit{ii}) screening and eligibility criteria (Section \ref{sec:screening_eligibility}). Figure \ref{fig:scheme} details the methods and material used. More in detail, this review aims to gather knowledge to answer the following research questions:

\begin{itemize}
    \item \textbf{RQ1}: Which solutions based on \textsc{llm}s are being developed (\textit{e.g.}, for assessment tasks)? (\textit{i.e.}, excluding multidisciplinary solutions that were not specifically intended for learning assistance)
    \item \textbf{RQ2}: Which educational solutions based on \textsc{llm}s involved students or teachers at any level of the development process (\textit{e.g.}, design, evaluation)?
\end{itemize}

\subsection{Data gathering}
\label{sec:data_gathering}

The data were extracted using Google Scholar\footnote{Available at \url{https://scholar.google.com}, April 2024.} with two search queries, specially designed to gather works within the educational field that leverage \textsc{llm}s: 

\begin{enumerate}
    \item \texttt{"education" AND "student" AND ("large language model" OR "GPT-3" OR "GPT-3.5" OR "GPT-4" OR "ChatGPT") -"review"}

    \item \texttt{"education" OR "student" AND ("large language model" OR "GPT-3" OR "GPT-3.5" OR "GPT-4" OR "ChatGPT") -"review"}
\end{enumerate}

Both queries have been restricted temporally since 2020, and the second query was applied to the title content exclusively. Note that duplicated elements and works that do not use \textsc{llm}s or do not indicate which model is exploited were not considered. The same applies to the works that assess the performance of \textsc{llm}s. In the end, 342 records were identified.

\subsection{Screening and eligibility criteria}
\label{sec:screening_eligibility}

This process was designed to identify works within the field of study that were written in English while at the same time discarding theoretical and review contributions (\textit{i.e.}, those that do not propose an \textsc{llm}-based solution but review existing solutions or hypothesize on the impact of \textsc{llm}s for educational purposes). The manual screening based on the above eligibility criteria resulted in 29 records. Note that this process distinguishes between published articles and conferences from pre-printed and non-peer-reviewed records. The criteria for selection and exclusion are presented in Table \ref{tab:criteria}.

\begin{table}[!htbp]
\centering
\footnotesize
\caption{\label{tab:criteria} Criteria for selection and exclusion.}
\begin{tabular}{ll} 
\toprule
\bf Criteria & \bf Description\\
\midrule
\multirow{2}{*}{Publication date} & Before 2020\\
& After 2020\\
\multirow{2}{*}{Writing language} & Not in English\\
& English \\
\multirow{2}{*}{Scope} & General\\
& Educational purpose\\
\multirow{2}{*}{Type of contribution} & Theoretical or review\\
& Virtual tutor proposal\\
\multirow{2}{*}{Language intelligence} & Not involving \textsc{llm}s\\
& Involving \textsc{llm}s\\
\bottomrule
\end{tabular}
\end{table}

\begin{figure}[!htbp]
\centering
\includegraphics[width=0.7\textwidth]{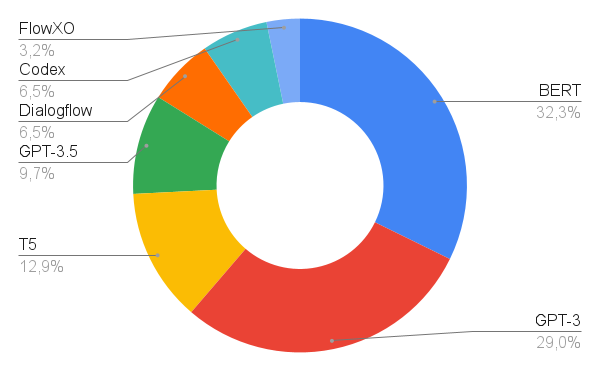}
\caption{\label{fig:models_chart}Distribution of the \textsc{llm}s in the records selected.}
\end{figure}

\begin{figure}[!htbp]
\centering
\includegraphics[width=0.7\textwidth]{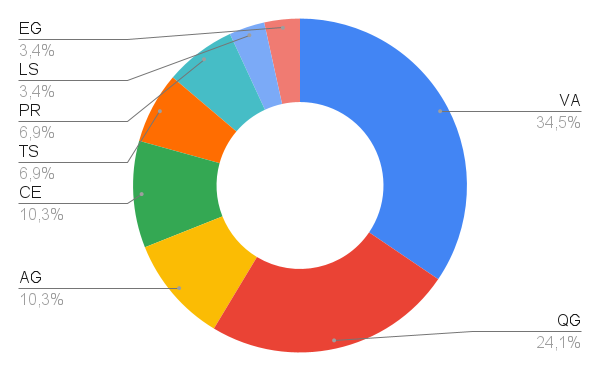}
\caption{\label{fig:tasks_chart}Distribution of the tasks in the records selected (\textsc{ag}: answer grading, \textsc{ce}: code explanation, \textsc{eg}: explanation generation, \textsc{ls}: learning software, \textsc{pr}: problem resolution, \textsc{qg}: question generation, \textsc{ts}: text summarization, \textsc{va}: virtual assistant).}
\end{figure}

Figure \ref{fig:models_chart} and Figure \ref{fig:tasks_chart} detail the distribution of the \textsc{llm}s used and applications in the works selected. Firstly, the most popular model is \textsc{bert}, followed by \textsc{gpt-3}, \textsc{t5}, and \textsc{gpt-3.5}. The low representativeness of the last \textsc{gpt} model contrasts with its popularity. The latter is due to the fact that the data gathering corresponds to the first quarter of 2023, that is, shortly after it was released. Thus, new works exploiting it are expected to be published shortly. Furthermore, the most common tasks these models perform in the selected works are as virtual assistants and question generation, as shown in Figure \ref{fig:tasks_chart}, followed by answer grading and code explanation/correction. Note that most works were published in 2022, with few records in 2021, showing a growth trend in 2023.

\section{Analysis \& discussion}
\label{sec:discussion}

Table \ref{tab:articles_conferences} lists the articles published in journals and the proceedings of conferences, taking into account their application, the model used, and code and data availability. Note that just the works by \cite{Liu2022,Mendoza2022,Tyen2022,zong2022solving, humphry2023potential,Nasution2023} provide enough information for reproducibility, while \cite{bhat2022towards,Essel2022, Mendoza2022,Moore2022,phillips2022exploring, yamaoka2022experience,Nasution2023} involved either teachers or students in the design or experimental plan.

\begin{table}[!htbp]
\centering
\footnotesize
\caption{\label{tab:articles_conferences} Selected articles published in journals or presented at conferences.}
\begin{tabular}{p{3cm}lp{1.7cm}l} 
\toprule
\bf \multirow{2}{*}{Application} & \bf \multirow{2}{*}{Authorship} & \bf \multirow{2}{*}{Model} & \bf Code/data\\
& & & \bf availability\\\midrule

\multirow{2}{*}{Answer grading} & \cite{ahmed2022application} & \textsc{bert} & \multirow{2}{*}{No}\\
& \cite{Moore2022} & \textsc{gpt-3}\\
\midrule

Code explanation & \cite{macneil2022generating} & \textsc{gpt-3} & No\\
\midrule

Explanation generation & \cite{humphry2023potential} & \textsc{gpt-3.5} & Yes\\
\midrule

Learning software & \cite{yamaoka2022experience} & \textsc{gpt-3} & No\\
\midrule

Problem resolution & \cite{zong2022solving} & \textsc{gpt-3} & Yes\\
\midrule

\multirow{4}{*}{Question generation} & \cite{bhat2022towards} & \textsc{gpt-3} \& \textsc{t5} & No\\
& \cite{Dijkstra2022} & \textsc{gpt-3} & No\\
& \cite{sharma2022generating} & \textsc{t5} & No\\
& \cite{Nasution2023} & \textsc{gpt-3.5} & Yes\\
\midrule

\multirow{3}{*}{Text summarization} & \cite{phillips2022exploring} & \textsc{gpt-3} & \multirow{3}{*}{No}\\
& \multirow{2}{*}{\cite{prihar2022identifying}} & \textsc{bert}, \textsc{sbert} \& Math\textsc{bert}\\
\midrule

\multirow{12}{*}{Virtual assistant} & \cite{Sophia2021} & Dialogflow & No\\

& \cite{BAHA2022} & Camem\textsc{bert} & No\\

& \cite{Calabrese2022} & De\textsc{bert}a & No\\

& \cite{Essel2022} & Flow\textsc{xo} & No\\

& \multirow{2}{*}{\cite{Liu2022}} & Ro\textsc{bert}a & \multirow{2}{*}{Yes}\\
& & Distill\textsc{bert}\\

& \cite{Mahajan2022} & Ro\textsc{bert}a & No\\

& \cite{Mendoza2022} & Dialogflow & Yes\\

& \cite{topsakal7framework} & \textsc{gpt-3.5} & No\\

& \cite{Tyen2022} & Ro\textsc{bert}a & Yes\\

& \multirow{2}{*}{\cite{designing_wang2022}} & \textsc{bert}, Al\textsc{bert} \& Distil\textsc{bert} & \multirow{2}{*}{No}\\
\bottomrule
\end{tabular}
\end{table}

Regarding answer grading applications, \cite{ahmed2022application} used the \textsc{bert} model. They exploited a modified version of the model based on triplets and the Siamese network, specially designed to generate sentences through semantically meaningful embeddings. The data set used is the one presented by \cite{Mohler2009}. The authors applied the question demoting technique as part of the preprocessing, thus removing from the answer those words also contained in the question. The authors performed the experiments with two different combinations of input data: (\textit{i}) the reference and student answers, and (\textit{ii}) the concatenation of the question and the reference answer, plus the answer provided by the student. Evaluation metrics include Pearson correlation coefficient (\textsc{pcc}) and root mean square error (\textsc{rmse}). The results are approximately 0.8 \textsc{pcc} and 0.7 \textsc{rmse}. \cite{Moore2022} presented another answer grading solution based on \textsc{gpt-3}. Unlike \cite{ahmed2022application}, the input data were gathered from an introductory chemistry course at the university level with almost 150 students. Moreover, the \textsc{gpt-3} model was trained with the \textsc{l}earning\textsc{q} data set \citep{Chen2018}, as in \cite{bhat2022towards}. Based on the assessment of the questions posed to experts in the chemistry field, the model was able to correctly evaluate \SI{32}{\percent} of the questions. 

Few works exist on code explanation and general explanation generation, learning software, and problem resolution. Firstly, \cite{macneil2022generating} proposed a \textsc{gpt-3}-based solution for code explanation based on 700 prompts. Note that it does not identify or correct errors. The main functionalities of the system encompass (\textit{i}) execution tracing, (\textit{ii}) identifying and explaining common bugs, and (\textit{iii}) output prediction. However, no results were provided. \cite{humphry2023potential} proposed a solution based on \textsc{gpt-3.5} to write conclusion statements about chemistry laboratory experiments. The evaluation of the solution relied on a discussion of features like readability and orthographic correctness of the generated text. Unlike the works above, which focused on textual input data, \cite{yamaoka2022experience} used the \textsc{gpt-3} model to exploit social media data, particularly from Instagram, for learning purposes. The proposed pipeline comprises (\textit{i}) detecting the relevant objects in the images, (\textit{ii}) extracting keywords to generate sentences related to those keywords, and (\textit{iii}) providing linguist information about the words that composed the sentence. The ultimate objective was to acquire new vocabulary. The experiments consisted of a small pilot study with three students from Osaka Metropolitan University. The only results reported were the average of unknown words, 2.2 in the generated sentences. Finally, \cite{zong2022solving} used \textsc{gpt-3} to identify and generate math problems involving systems of two linear equations. The experiments consisted of (\textit{i}) problem classification into five categories, (\textit{ii}) equation extraction from word problems, and (\textit{iii}) generation of similar exercises. The authors prepared the input data \textit{ad hoc}. The accuracy of the results obtained in each of the three tasks above was \SI{75}{\percent} (averaging the five categories); \SI{80}{\percent} (with fine-tuning), and \SI{60}{\percent} (also averaging the five categories), respectively.

Regarding question generation, several representative examples were found in the literature. \cite{bhat2022towards} used both \textsc{gpt-3} and \textsc{t5} models, \textsc{gpt-3} for question generation combined with a concept hierarchy extraction model, and \textsc{t5} for the evaluation in terms of learning usefulness of the generated questions. The input data consisted of textual learning materials from a university data science course. More in detail, the concept hierarchy extraction method exploited the \textsc{moocc}ube\textsc{x} pipeline \citep{Yu2021}, which extracts key concepts following a semi-supervised approach. Note that evaluation also involved computing the information score metric and manual assessment by human annotators. The experimental results obtained with the \textsc{l}earning\textsc{q} data set \citep{Chen2018} show that almost \SI{75}{\percent} of the generated questions were considered useful by the \textsc{gpt-3} model, with an agreement slightly higher than \SI{65}{\percent} when compared to manual evaluation. Similarly, \cite{Dijkstra2022} created EduQuiz with \textsc{gpt-3}, a multi-choice quiz generator for reading comprehension exploiting the \textsc{eqg-race} data set\footnote{Available at \url{https://github.com/jemmryx/EQG-RACE}, April 2024.} \citep{Jia2021}. The authors evaluated the performance of EduQuiz using standard metrics, \textsc{bleu-4}, \textsc{rouge-l}, and \textsc{meteor}. Results attained 36.11, 11.61, and 25.42 for these metrics, respectively. Additionally, \cite{sharma2022generating} proposed a fine-tuning pipeline composed of context recognition and paraphrasing, filtering irrelevant output, and translation to other languages for question generation at different levels using the \textsc{t5} model. The authors used the data set by \cite{Mohler2011} (an updated version of the data set used in \cite{ahmed2022application}). The evaluation metrics computed were \textsc{blue} \citep{Papineni2002} and \textsc{meteor} \citep{Lavie2007}. The results for the two metrics above were 0.52 and 57.66, respectively. Thus, compared with the question generation solution by \cite{Dijkstra2022}, \cite{sharma2022generating} obtained a more competitive \textsc{meteor} value. Ultimately, \cite{Nasution2023} used \textsc{gpt-3.5} for question generation. To assess the generated questions' reliability or internal consistency, the Cronbach's alpha coefficient \citep{Taber2018} was computed, resulting in 0.65. Answers from a survey performed to almost 300 students show that \SI{79}{\percent} of the generated questions were relevant, \SI{72}{\percent} were moderately clear, and \SI{71}{\percent} were of enough depth.

In contrast, \cite{phillips2022exploring} used \textsc{gpt-3} to create summaries of students' chats in collaborative learning. Moreover, this solution detected confusion and frustration in the student's utterances. Input data was gathered from secondary school students in an ecosystem game. The authors briefly discussed how the system could provide advantageous knowledge to teachers about their interaction in a collaborative learning environment, but no further analysis or results were provided. Conversely, \cite{prihar2022identifying} proposed a learning assistant based on the \textsc{bert} model and its variations (\textit{i.e.}, \textsc{sbert} and Math\textsc{bert}) to generate support messages from chat logs obtained from fundamental interactions between a live \textsc{up}chieve tutor available at the \textsc{assist}ments learning platform and the students. Even though \SI{75}{\percent} of the generated messages were identified as relevant by manual human evaluation, these messages had a negative impact on the student's learning process, as the authors explained.

The most common application uses \textsc{llm}s as virtual assistants. \cite{Sophia2021} created \textsc{edubot} exploiting Dialogflow. Its main limitation lies in the basic language understanding capabilities (\textit{i.e.}, low variability in the responses provided), particularly regarding the user's emotions. \cite{BAHA2022} developed Edu-Chatbot exploiting the Xatkit framework. The system comprises an encoder based on Camem\textsc{bert} and a decoding module for student intent recognition. Unfortunately, the intent classification decoder is based on a pre-defined set of recognized actions (\textit{e.g.}, simple questions, animations, videos, and quizzes). Thus, the language intelligence of the solution is limited. Furthermore, no evaluation was performed. \cite{Calabrese2022} presented a virtual assistant prototype for Massive Online Open Courses (\textsc{mooc}s). Their objective was to reduce the teaching load and maintain the quality of learning. Thus, its architecture allows the teacher to intervene in those questions that have not been resolved satisfactorily. More in detail, they used a personalized version of \textsc{bert}. The questions answered by the teacher are included in an additional document and allow the \textsc{bert} model to be improved. In contrast, \cite{Essel2022} involved 68 undergraduate students in evaluating the solution developed using Flow\textsc{xo} and integrated into WhatsApp. Qualitative evaluation on the end-user's preferences of the virtual assistant instead of traditional interaction approaches with the teachers reached \SI{58.8}{\percent}. Additionally, \cite{Liu2022} presented a virtual assistant for online courses to resolve general and repetitive doubts about content and teaching materials. This system incorporates a sentiment analysis module to analyze the response's satisfaction based on the student's dialogue. They used two fine-tuning versions of \textsc{bert} model with an accuracy of \SI{82}{\percent} and \SI{90}{\percent} for the correct detection of the content and student's sentiment, respectively. Moreover, \cite{Mahajan2022} created a system for students to improve their knowledge of the English language that allows them to obtain information on the meaning of words, make translations, resolve pronunciation doubts, etc. The authors exploited the Ro\textsc{bert}a model with an accuracy greater than \SI{98}{\percent} in communication intent detection. Similarly, \cite{Mendoza2022} created a virtual assistant intended for academic and administrative tasks but exploiting Dialogflow\footnote{Available at \url{https://dialogflow.cloud.google.com}, April 2024.}. The Cronbach’s alpha coefficients during the evaluation of the system exceeded 0.7. In contrast, \cite{topsakal7framework} presented a foreign language virtual assistant based on the \textsc{gpt-3.5} model combined with augmented reality. The authors claim that this combination attracted students' attention and motivated them through entertaining learning thanks to gamification. In this case, the language model was used to establish a dialogue with the end users. Unfortunately, no results were discussed. Moreover, \cite{Tyen2022} proposed a virtual assistant for second language learning with difficulty level adjustment in the decoder module and evaluated by experienced teachers. The system exploits Ro\textsc{bert}a fine-tuned with a Cambridge exams data set. The system attained Spearman and Pearson coefficients of 0.755 and 0.731, respectively. Finally, \cite{designing_wang2022} developed an educational domain-specific chatbot. Its goal is to reduce pressure on teachers in virtual environments and improve response times by easing communication between students and teachers. They used Natural Language Understanding (\textsc{nlu}) techniques on variations of the \textsc{bert} model for the classification of intents and response generation. It presented an accuracy of \SI{88}{\percent} in detecting intents. However, its values are lower than \SI{50}{\percent} regarding semantic analysis.

Table \ref{tab:pre-printed} lists the selected pre-printed or non-peer-reviewed works, taking into account their application, model used, and reproducibility feature. In this case, \cite{daSilva2022}\footnote{\label{da_silva}Available at \url{https://arc.cct.ie/cgi/viewcontent.cgi?article=1026&context=ict}, April 2024.}, \cite{Zhang2022}\footnote{\label{zhang}Available at \url{https://arxiv.org/pdf/2209.14876.pdf}, April 2024.} and \cite{Christ2023}\footnote{\label{christ}Available at \url{https://fbmn.h-da.de/fileadmin/Dokumente/Studium/DS/WS2022_MDS_Thesis_Paul_Christ_THE.pdf}, April 2024.} provide enough information for reproducibility, while \cite{Zhang2022}\footnoteref{zhang} involved either teachers or students in the design or experimental plan.

\begin{table}[!htbp]
\centering
\footnotesize
\caption{\label{tab:pre-printed} Selected pre-printed or non peer-reviewed records.}
\begin{tabular}{p{3.5cm}lll} 
\toprule
\bf \multirow{2}{*}{Application} & \bf \multirow{2}{*}{Authorship} & \bf \multirow{2}{*}{Model} & \bf Code/data\\
& & & \bf availability\\\midrule

Answer grading & \cite{Hardy2021}\footnoteref{hardy} & \textsc{sbert} & No \\
\midrule

\multirow{2}{*}{Code correction} & \cite{Zhang2022}\footnoteref{zhang} & \multirow{2}{*}{Codex} & Yes \\
& \cite{Phung2023}\footnoteref{phung} & & No \\
\midrule

Problem resolution & \cite{Cobbe2021}\footnoteref{cobee} & \textsc{gpt-3} & No\\
\midrule

\multirow{3}{*}{Question generation} & \cite{daSilva2022}\footnoteref{da_silva} & \textsc{t5} & Yes \\
& \cite{Raina2022}\footnoteref{raina} & \textsc{gpt-3} \& \textsc{t5} & No \\
& \cite{Christ2023}\footnoteref{christ} & Distil\textsc{bert} & Yes\\

\bottomrule

\end{tabular}

\end{table}

The distribution of applications is similar to the peer-reviewed records. \cite{Hardy2021}\footnote{\label{hardy}Available at \url{https://arxiv.org/ftp/arxiv/papers/2112/2112.11973.pdf}, April 2024.} developed an automatic evaluation system for reading and writing exercises. The system uses the \textsc{sbert} model, among others, to capture semantic data and provide valuable insights related to the student's skills, using \textsc{asap-aes}\footnote{Available at \url{https://www.kaggle.com/c/asap-aes}, April 2024.} and \textsc{asap-sas}\footnote{Available at \url{https://www.kaggle.com/c/asap-sas}, April 2024.} data sets. Particularly, they exploited the passage-dependent sentence-\textsc{bert} model trained using curricular learning \citep{Graves2016}. The results from the Quadratic Weighted Kappa (\textsc{qwk}) metric reached 0.76 on average.

Regarding code correction, \cite{Zhang2022}\footnoteref{zhang} presented \textsc{mmapr}, an error identification and correction system for code development based on the Open\textsc{ai} Codex model\footnote{Available at \url{https://openai.com/blog/openai-codex}, April 2024.}. The system fixes semantic and syntax errors by combining iterative querying, multi-modal prompts, program chunking, and test-case-based selection of a few shots. Results obtained with almost 300 students reached \SI{96.50}{\percent} in corrected code rate with the few-shots-based approach. \cite{Phung2023}\footnote{\label{phung}Available at \url{https://arxiv.org/pdf/2302.04662.pdf}, April 2024.} presented a similar solution to \textsc{mmapr} for code correction named \textsc{p}y\textsc{f}i\textsc{xv}. The main difference is that the Codex model, combined with prompt engineering, explains the detected errors. Moreover, the explanations are also validated in terms of suitability for the students. The system has been tested with TigerJython \citep{Kohn2020} and Codeforces\footnote{Available at \url{https://codeforces.com}, April 2024.} data sets. The precision attained \SI{76}{\percent} in the most favorable scenario with the TigerJython data set.

\cite{Cobbe2021}\footnote{\label{cobee}Available at \url{https://arxiv.org/pdf/2110.14168.pdf?curius=520}, April 2024.} elaborated a data set of 8.5\textsc{k} elementary school mathematical problems called \textsc{gsm8k}. Then, the \textsc{gpt-3} model was used to generate comprehensible explanations of these problems, combining natural language and mathematical expressions. The authors trained verifiers to enhance the performance of the model beyond fine-tuning. Ultimately, they concluded that this approach enhanced the overall performance.
 
Subsequently, \cite{daSilva2022}\footnoteref{da_silva} developed an automatic questionnaire generation system using the \textsc{t5} model and applying the fine-tuning technique, named \textsc{querai}. The \textsc{t5} model was evaluated with Skip Thought Vectors (\textsc{stv}), Embedding Average Cosine Similarity (\textsc{eacs}), Vector Extrema Cosine Similarity (\textsc{vecs}), and Greedy Matching Score (\textsc{gms}) metrics, with results higher than 0.8 except for \textsc{vecs}. Summing up, the accuracy of the pay-per-subscription solution is \SI{91}{\percent}. Similar to \cite{daSilva2022}\footnoteref{da_silva}, \cite{Raina2022}\footnote{\label{raina} Available at \url{https://arxiv.org/pdf/2209.11830.pdf}, April 2024.} developed a multiple-choice question generation solution to generate both questions and the set of possible answers using apart from \textsc{t5}, the \textsc{gpt-3} model, both trained with the \textsc{race++} \citep{Liang2019} data set composed of middle, high and college level questions. The results obtained are similar between the two models with an accuracy of \SI{80}{\percent} (11 percentage points lower than the solution by \cite{daSilva2022}\footnoteref{da_silva}). Note that the authors also measured the number of grammatical errors and other features like diversity and complexity. The lowest values are related to the diversity of the questions generated. In the best scenario, the \textsc{t5} model attained \SI{60}{\percent} accuracy, approximately. More recently, \cite{Christ2023}\footnoteref{christ} used \textsc{bert} to generate \textsc{sql}-Query exercises automatically. Experiments with knowledge graphs and natural language building were also performed as a baseline. The authors concluded that the Distil\textsc{bert}-based approach generates descriptions that are, on average, almost \SI{50}{\percent} shorter and with a \SI{20}{\percent} decrease in term frequency compared to the \textsc{nlp} baseline. 

\textsc{gpt-3} and the different adaptations of the \textsc{bert} model are the most popular alternatives in the sample regarding answer grading, code explanation and general explanation generation, learning software, problem resolution, question generation, and text summarization. When it comes to their use as virtual assistants, the variety of models used increases. The current lower costs of the \textsc{gpt-3.5} model will motivate a rapid increase in its use in the coming years. However, \textsc{bert} robustness as an entity detector, with evaluation metrics above \SI{80}{\percent} in several of the discussed works, made it a reference for developing educational software tools. Unfortunately, most works reviewed do not provide the code or data used for their analysis, making reproducibility difficult. Finally, regarding the risks of exploiting \textsc{llm}s for educational tasks, they are transversal (\textit{e.g.}, for automatic question generation and as virtual assistants, the two most popular applications identified). The lack of transparency of the models (\textit{i.e.}, the rationale behind their functioning, such as difficulty adjustment in question generation) could negatively impact the end-users. Regarding their use as virtual tutors, reinforcement learning from human feedback is essential to gain control over their operation and ensure fairness. Ultimately, the risk of poor accuracy must be palliated by including the probabilistic confidence of their response.

\section{Conclusion}
\label{sec:conclusion}

\textsc{llm}s represent an undeniably mega-trend in the current century in many fields and industrial sectors. In the particular case of learning, these generative \textsc{ai}-based solutions have produced a considerable buzz. Accordingly, they enable hands-on learning and are commonly used in classrooms nowadays. Compared to previous \textsc{ai} solutions and traditional methodologies, which focused primarily on modifying the textual input, advanced \textsc{llm}s can generate on-the-fly human-like utterances, enhancing pedagogical practice and providing personalized assessment and tutoring.

Given the popularity of \textsc{llm}s, this work is the first to contribute with a comprehensive overview of their application within the educational field. Paying particular attention to those that involved students or teachers in the design or experimental plan. From the 342 records obtained during data gathering, 29 works passed the screening stage by meeting the eligibility criteria. They were discussed, taking into account their application within the educational field, the model used, and code and data availability features. Results show that the most common tasks performed as virtual assistants are question generation, answer grading, and code correction and explanation. Moreover, the most popular model continues to be \textsc{bert}, followed by \textsc{gpt-3}, \textsc{t5}, and \textsc{gpt-3.5} models. In the end, this review identified 9 reproducible works and 8 solutions that involved either teachers or students in the design or experimental plan.

Due to the recent launch of the \textsc{gpt-4} model within the Chat\textsc{gpt} application, new works are expected to be published soon and will be analyzed as part of future work. Moreover, as future work, we will study the ethical implications of \textsc{llm}s (\textit{i.e.}, their transparency and fairness behavior caused by the training data and privacy) and how the solutions discussed can be integrated into the education curricula, as well as their shortcomings and risk to academic integrity (\textit{e.g.}, plagiarism concerns). Finally, attention will be paid to those works that propose innovative teaching practices with \textsc{llm}s and explore the use of \textit{ad hoc} solutions through personal language models in the field.

\section*{Declarations}

\subsection*{Funding}

This work was partially supported by: (\textit{i}) Xunta de Galicia grants ED481B-2021-118 and ED481B-2022-093, Spain; and (\textit{iii}) University of Vigo/CISUG for open access charge.

\subsection*{Competing interests}
The authors have no competing interests to declare relevant to this article's content.

\subsection*{Ethics approval}

Not applicable

\subsection*{Consent to participate}

Not Applicable

\subsection*{Consent for publication}

Not Applicable

\subsection*{Availability of data and material}

The used data is openly available.

\subsection*{Authors' contributions}

\textbf{Silvia García-Méndez}: Conceptualization, Methodology, Software, Validation, Formal analysis, Investigation, Resources, Data Curation, Writing - Original Draft, Writing - Review \& Editing, Visualization, Supervision, Project administration, Funding acquisition. \textbf{Francisco de Arriba-Pérez}: Conceptualization, Methodology, Software, Validation, Formal analysis, Investigation, Resources, Data Curation, Writing - Original Draft, Writing - Review \& Editing, Visualization, Supervision, Project administration, Funding acquisition. \textbf{Carmen Somoza-López}: Conceptualization, Writing - Review \& Editing.

\bibliography{mybibfile}

\end{document}